\documentclass{jmlr}% new name PMLR (Proceedings of Machine Learning)

\usepackage{longtable}% for long tables
\usepackage{multicol}% for long tables
\usepackage{booktabs}
\usepackage{todonotes}
\usepackage[load-configurations=version-1]{siunitx} % newer version
\theorembodyfont{\upshape}
\theoremheaderfont{\scshape}
\theorempostheader{:}
\theoremsep{\newline}

\jmlrvolume{1}
\jmlryear{2018}
\jmlrworkshop{AutoML 2018}

\title{Towards AutoML in the presence  of Drift: first results %Towards Autonomous Lifelong Machine Learning with Drift%\titlebreak This Title Has
}

  \author{\Name{Jorge G. Madrid} \Email{jorgegus.93@gmail.com}\\
\addr INAOE, Mexico\\
   \Name{Hugo Jair Escalante} \Email{hugojair@inaoep.mx}\\
\addr INAOE, Mexico\\
   \Name{Eduardo F. Morales} \Email{emorales@inaoep.mx}\\
\addr INAOE, Mexico\\
   \Name{Wei-Wei Tu} \Email{tuwwcn@gmail.com}\\
\addr 4Paradigm Inc., China \\
\Name{Yang Yu} \Email{yuy@nju.edu.cn}\\
\addr Nanjing University, China\\
\Name{Lisheng Sun-Hosoya} \Email{cecile829@gmail.com}\\
\addr UPSud, U. Paris-Saclay, France \\
\Name{Isabelle Guyon} \Email{guyon@chalearn.org}\\
\addr UPSud/INRIA, U. Paris-Saclay, France and ChaLearn, USA \\
\Name{Michele Sebag} \Email{Michele.Sebag@lri.fr}\\
\addr CNRS, U. Paris-Saclay, France}
\begin{document}

\maketitle

\begin{abstract}
%\todo[author=HJ,inline]{Suggestions for the title are welcome}
%In many real-world machine learning applications, AutoML is strongly needed due to the limited machine learning expertise of developers. 
Research progress in AutoML has lead to state of the art solutions that can cope quite well with supervised learning task, e.g., classification with AutoSklearn. However, so far these systems do not take into account the changing nature of evolving data over time (i.e., they still assume i.i.d. data); even when this sort of domains are increasingly available in real applications (e.g., spam filtering, user preferences, etc.). We describe a first attempt to develop an AutoML solution for scenarios in which data distribution changes relatively slowly over time and in which the problem is approached in a lifelong learning setting. We extend Auto-Sklearn with sound and intuitive mechanisms that allow it to cope with this sort of problems. The extended Auto-Sklearn is combined with concept drift detection techniques that allow it  to automatically determine when the initial models have to be adapted.  We report experimental results in benchmark data from AutoML competitions that adhere to this scenario. Results demonstrate the effectiveness of the proposed methodology. 
%deal with data distribution changes using a modification to a robust solution (auto-sklearn). A global replacement and a model management strategy are described and tested with experiments in three real and one synthetic datasets.
%Research progress in machine learning has lead to very effective learning machines able to cope decently with a vast number of problems coming from heterogeneous domains and comprising a wide diversity of variables and tasks (e.g., regression, clustering, classification). Because the number of available algorithms and their proved performance, the current problem seems to be how to choose the right algorithm for a given task and how to set its hyperparameters. AutoML is the machine learning field that aims at automatically solving machine learning tasks without human supervision. Recent progress in AutoML has lead to very effective methods in this setting, however, so far the considered scenario is limited in the sens that only the standard AutoML setting has been studied. In this paper we 

\end{abstract}

\section{Introduction}
\label{sec:intro}
Autonomous Machine Learning (AutoML) is the field focusing on methods that aim at automating different stages of the machine learning process. AutoML solutions are increasingly receiving more attention from both the ML community and users because of (1) the large amounts of data readily available everywhere, and (2) the lack of domain and/or ML experts who can advise/supervise the development of ML-based systems.

Although progress in AutoML is vast, the considered scenarios are somewhat constrained, e.g., in the type of approached problem, in the assumptions on data, in the size of datasets, etc. In this context, one of the most desirable features for AutoML methods is  to work under a \emph{lifelong  machine learning} (LML) setting. LML refers to systems that can sequentially learn many tasks from one or more domains in its lifetime~\citep{danny13}, these systems (not restricted to supervised learning) require the ability to retain knowledge, adapt to changes and transfer knowledge when learning a new task. An AutoML method that learns from different tasks and that is able to adapt itself during its lifetime would comprise a competitive and robust  all-problem machine learning solution.  

This paper aims at exploring the viability of AutoML methods to operate in a LML setting, in particular in a scenario where the targets evolve over time, that is, in the presence of concept drift. We modify the Auto-Sklearn method with mechanisms that allow it to deal with the drift phenomenon in a simplified LML evaluation scenario. %\footnote{Please note that we adopted a reductionist  LML scenario for evaluation. }. 
The proposed mechanisms are sound and highly intuitive, yet very useful. We perform experiments in benchmark data from concept drift and AutoML with drift. % past and ongoing AutoML competitions, one of them focusing in AutoML in LML with drift\footnote{\url{https://www.4paradigm.com/competition/nips2018}}. Additionally, we also evaluate the proposed method in benchmark datasets with concept drift. 
Experimental results reveal that the proposed mechanisms allow Auto-Sklearn to successfully  cope with drift. To the best of our knowledge these are the first results reported on AutoML for a simplified LML scenario in the presence of concept drift\footnote{One should note that our work is different from standard data-stream / concept drift learning methods in the sense that in AutoML we aim to autonomously find preprocessing, feature selection and classification methods (together with hyper parameter optimization), while in most of the existing literature the problem is reduced to modify the classification model.   }.  %\footnote{One should note, however, that there there is a whole research field focusing on concept drift and data stream classification. Yet, this proposal is focused on AutoML}. 

%The remainder of this paper is organized as follows. Next section provides a review of related work on AutoML, LML and concept drift. Section~\ref{sec:automlll} describes the considered scenario and the proposed extensions to Auto-Sklearn. Next, Section~\ref{sec:experiments} describes experiments and presents results. Finally, Section~\ref{sec:conlcusions} outlines preliminary conclusions of this work. 
%A first step towards LML is presented in this work, however, the problem is limited to adapting an existing robust AutoML system to deal with changes in data distributions over time, specifically, the Auto-sklearn software~\cite{autosklearn} was modified developing a schema to adapt a classification model to drifting concepts, with it some simple solutions are proposed and compared, finally, some limitations of the solutions are described.
%As a natural evolution of these challenges, we are organizing a 
%In this proposal, we aim to explore areas of AutoML that have not been studied so far, and that are present in almost every possible application of AutoML. As previously mentioned, the novel components of the proposed challenge are: the use of \textbf{large scale} datasets coming from \textbf{real-world applications}, where data are subject to the \textbf{concept drift} phenomenon and from adopting a \textbf{lifelong} evaluation. Accordingly, we will provide data and an evaluation framework that allow us to draw interesting conclusions and make further progress on AutoML.  

\section{Related work}
\label{sec:rw}
%\textcolor{red}{Lisheng, Isabelle, Michele?}
% Formal background: key concepts

% * stationary distribution 

% * one example at a time (incremental learning = online learning); or one dataset at a time (domain adaptation)

% * data streaming: one batch at a time; distribution can evolve; change detection in data mining; G. Hulten, L. Spencer, and P. Domingos. Mining time-changing data streams. In KDD 2001.

% * covariate shift P(x) non stationnaire; P(y|x) cst.
%  Machine Learning in Non-Stationary Environments
% Introduction to Covariate Shift Adaptation; Masashi Sugiyama and Motoaki Kawanabe. (\url{https://books.google.fr/books?id=Akd-8xFKac4C&pg=PA19&source=gbs_toc_r&cad=3#v=onepage&q&f=false})

Although the term `AutoML' was coined recently by F. Hutter and collaborators ~\citep{autoweka,hutter2009automated}, the problem of hyper-parameter selection has been studied for several decades in the machine learning community~\citep{bozdogan1987model,bengio2000gradient,bergstra2011algorithms,bergstra2012random}. However, the original emphasis was on over-fitting avoidance while, with the emergence of ``big data'' the current emphasis is on search efficiency. Novel effective approaches have been proposed in the academic literature that have become wide spread among practitioners because they are both theoretically well founded and practically efficient: (1) Bayesian Optimization (BO)  methods build a posterior $p(model\vert data)$ by applying candidate models to the input data and use this posterior distribution to guide the search (e.g.~\citep{hutter2011sequential,swersky2014freeze}). (2) Complementary to BO, Meta Learning develops a set of meta-features capturing the nature of data, which are then used to infer the model performance based on past experiences on similar data, without actually training the model (e.g.~\citep{munoz2018instance}). (3) Evolutionary Algorithms (EA) learn a distribution over hyper-parameters and updates it to help the search (e.g.~\citep{real2017large}). (4) Reinforcement learning approaches (RL), where the hyper-parameter optimization problem is formulated as learning an efficient policy to move in the hyper-parameter space and solved using RL techniques (e.g.~\citep{zoph2016neural,baker2016designing}). The algorithm Auto-sklearn used in this paper is based on a BO approach initialized with Meta Learning. \par

Hyper-parameter selection usually focuses on solving tasks in isolation, but, if tasks show some sequential dependency, a lifelong learning (LML) approach can be applied to continuously retain and immigrate knowledge across tasks and make the future learning more efficient. The concept of LML was first introduced in robot control~\citep{thrun1995lifelong}, where, inspired by the fact that a robot might be exposed to various learning tasks during their lifetime and the knowledge learned might be transfered to make future learning more efficient, they proposed EBNN (explanation-based neural network): suppose the robot perform all tasks in the same environment (e.g. housekeeping robot), then its knowledge about the environment (called \textit{action model}, the mapping between state-action pairs to the next state) can be transfered to new task using back-propagation. Since then, diverse techniques have been developed in lifelong supervised learning ~\citep{silver2002task, silver1996parallel,silver2001selective, chen2018lifelong, fei2016learning}.

% Daniel L. Silver et al. proposed Task Rehearsal Methods (TRM)~\citep{silver2002task}, for learning tasks suffering from an impoverished training set by augmenting it with virtual examples generated from the representation of formerly successfully learned tasks. The relatedness among tasks in TRM is dealt with the $\eta$MTL network~\citep{silver1996parallel,silver2001selective}. \cite{chen2018lifelong} developed a lifelong learning approach for sentiment classification, where the problem is solved in a Naive Bayes framework, and the knowledge transfer is done by adding penalty terms to the optimization formulation. Fei et al.~\citep{fei2016learning} broadened the concept and proposed \textit{cumulative learning} where the built learner should be able to incorporate learned knowledge to detect data from unseen classes in test time. \par

% \subsection{Concept drift}
Due to the big volume of data being constantly generated, many prediction problems need a model that continuously receives data and thus cannot work in an off-line/static mode with historical data. Some of these environments are non-stationary where data distributions change over time, this phenomenon is known as concept drift~\citep{schlimmer1986beyond,widmer1996learning, gama}. Numerous research work have been done for adapting models to data streams with presence of concept drift. \cite{hulten2001mining} developed \textsc{CVFDT} where a Decision Tree is maintained up-to-date with regard to a sliding window of examples: an alternative sub-tree is built when drift is detected in the window, and is used to replace the old sub-tree when this new one becomes more performing. Then, by proving the out-performance of ensembling compared to single classifiers in the concept drift environment, \cite{wang2003mining} proposed weighted classifier ensemble, where adaptation is done by dynamically updating the weight for each base classifier according to its expected prediction accuracy on current test stream. To address the problem of sliding window size, \cite{bifet2007learning} introduced \textsc{Adwin} in which the window is resized based on changing rate observed in the window, \textsc{Adwin} is also used widely as a change detector as in \cite{bifet2010leveraging} and \cite{van2014algorithm}. The same authors then came up with \textsc{Leverage Bagging} (\cite{bifet2010leveraging}) where examples are weighted by a well parameterized Poisson distribution to add randomization in input data that improve the accuracy of ensembling. \cite{van2014algorithm}, \cite{rossi2014metastream} and \cite{van2015having} collaborated meta-learning techniques to solve the algorithm selection problem: predict best model for next upcoming stream based on meta-knowledge collected from previous streams. 

Although work for processing streams and data in the presence of drift is vast, to the best of our knowledge existing work as not approaching the AutoML setting: automatically building and updating a \emph{full model}~\citep{psms}, that is a model that comprises data preprocessing, feature selection and classification model and that optimizes hyperparameters of the whole model. 

In the scenario of LML-AutoML, the  learning algorithm should be able to incorporate new data and update previous predictive models. In this paper, we follow the 3 steps of on-line adaptive learning procedure proposed in~\citep{gama}: (1)\textit{Predict}. When new example $X_t$ arrives, a prediction $\hat{y}ˆt$ is made using the current
model $L_t$. (2)\textit{Diagnose}. After some time the true label $y_t$ is received and the loss can be estimated as
$f(\hat{y}_t, y_t)$, and (3)\textit{Update}. use the example $(X_t, y_t)$ to update the model to obtain $L_{t+1}$. Several methods that cope with concept drift use an explicit drift detection algorithm~\citep{awe,tbe,sea}. One of the most used detector which is used as a benchmark in new detection methods is the Drift Detection Method (DDM)~\citep{ddm}. It is a method that controls the trace of the on-line error by modeling the classification errors with a Binomial distribution. Some extensions to DDM have been developed to improve its performance. Fast Hoeffding Drift Detection Method (FHDDM)~\citep{fhddm} uses a sliding window and Hoeffding's inequality to detect drifting points earlier, experimental results also show that the number of false positive and false negatives is minor~\citep{fhddm,tornado}. Other methods extend FHDDM, however, these require additional parameters.
%This work focuses on real concept drift that happens when the distribution of the classes changes, it is different from virtual concept drift which refers to the changes of the distribution of features without affecting the resulting class. It also assumes that \textit{feedback} is received after making a prediction making possible the diagnose phase in the adaptive learning process.

\section{AutoML scenario and proposed methods}
\label{sec:automlll}
%\textcolor{red}{All, lead by Jorge, Eduardo, Hugo}

\begin{figure}[htb]
\begin{center}
\includegraphics[width=0.5\linewidth]{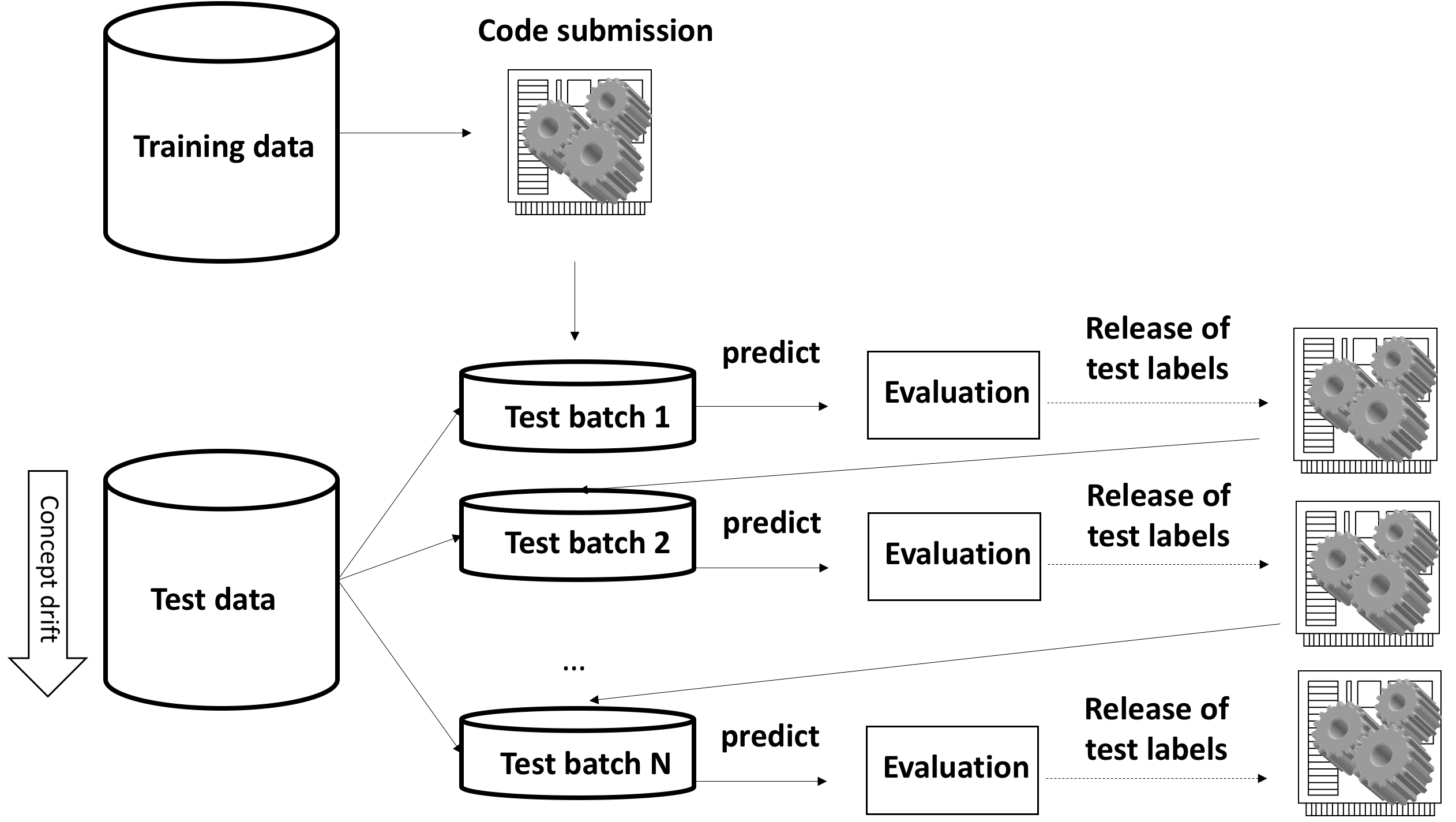}
\caption{Evaluation scenario considered in the AutoML3 challenge.}
\label{fig:scenario}
\end{center}
\end{figure}

As previously mentioned we consider a simplified LML - AutoML with a concept drift scenario, which  is precisely the scenario considered in the forthcoming  AutoML3\footnote{\url{https://www.4paradigm.com/competition/nips2018}} challenge. The aim is assessing the robustness of methods to concept drift and its lifelong learning capabilities. It is assumed that there is an initial training  (labeled) dataset available, and multiple batches of test data. Also, we assume that data are temporally dependent and are subject to an underlying form of the concept drift phenomenon.  For the LML - AutoML evaluation, datasets will be split into sequential batches so that the lifelong scenarios can be evaluated: the test data will come in the form of sequential batches of data, each batch needs to be predicted at first, then the target values will be revealed and thus become a batch of new training data. From the model predictions and the revealed target values of a specific test batch, its performance can be evaluated. The performance of the model is given by taking the average across batches.  Figure~\ref{fig:scenario} illustrates the considered scenario. 

\subsection{Auto-sklearn}
%\todo[author=HJ,inline]{We should reduce this}
Auto-sklearn is an AutoML solution that has succeed in recent academic competitions~\citep{autosklearn,DBLP:conf/icml/GuyonCEEJLMRRSS16,DBLP:conf/ijcnn/2015}. In our opinion, it % To the best of our knowledge, this 
is THE state of the art on AutoML  and for that reason we considered it for our study. Auto-Skelearn is implemented in scikit-learn~\citep{scikit}, it initially comprised 15 classification algorithms, 14 preprocessing methods, and 4 data preprocessing methods.  Similarly to Auto-WEKA~\citep{autoweka}, Feurer et al. approach AutoML as a Combined Algorithm Selection and Hyper-parameter (CASH) or full model selection problem~\citep{autosklearn,psms}, which they tackle using SMAC, a three-based Bayesian optimization method. There are two key components that make Auto-Sklearn so competitive. 
%There are two improvements to this AutoML approach that have made Auto-Sklearn a popular solution and that has led it to win multiple challenges including the  ChaLearn 2015-2016 AutoML and the AutoML 2018 challenges. 
The first is based on \textit{meta-learning}, complementary to Bayesian optimization, it is used to warmstart the optimization by quickly suggesting instantiations of a framework that are likely to perform well. The meta-learning was done in an off-line phase where 38 meta-features were learned from 140 OpenML datasets~\citep{openml}.  The second feature is the automated ensemble construction of models evaluated during optimization, when finding the best model instead of discarding the rest of the models found in the Bayesian optimization process, Feurer, et. al. store them and then build and ensemble using a greedy ~\textit{ensemble selection} algorithm. %Both improvements are considered in the solutions for adapting an ensemble classifier to drifting concepts proposed in this work.

\subsection{Proposed method}
Auto-Sklearn was modified with basic mechanisms for allowing it to cope with incoming data as depicted in Figure~\ref{fig:scenario}. The overall proposed procedure is shown in Algorithm~\ref{algo:1}. 
With a first batch, an initial ensemble model is learned by using Auto-Sklearn. With the generated model predictions are made for the next batch. After the predictions are made, feedback for this batch is received (i.e., ground truth), using a drift detector these predictions are diagnosed to determine if drift has happened. %\todo[author=HJ,inline]{@Jorge: the detection is performed after predictions are made?, please clarify} 
If the detector triggers the drift alarm, the current ensemble model is adapted by using one of the mechanisms described below. After this, the system receives the next batch and the process is repeated until no batches are left. In the rest of this section we describe the components of the proposed method, namely: the drift detector and the proposed adaptation mechanisms. 
\begin{algorithm}\small{
	\KwData{$D(X,y)$ examples}
    Take a batch $D'_t$ of size n, $D'_t(X',y') \in D(X,y)$;\;
    $T_t \leftarrow$ learn a model with auto-sklearn using $D'_t$\;
    \While{there is data in D}
    {
    	Take next batch $D'_{t+1}$;\;
        $\hat{y} \leftarrow$ Make predictions with $T_t$;\;
        \For{$y_j \in \hat{y}$}
        {
        	drift\_detected = Detector($y_j == y'_j \in y'$)\; //drift will be detected with model performance
        }
        \eIf{drif\_detected}
        {
        	$T_{t+1} = adapt(T,D'_{t+1})$;\;
            Detector.reset();\;
            $t=t+1$
        }
        {
        	$T_{t+1} = T_{t}$;\;
            $t=t+1$
        }
    }
    }
\caption{Drift adaption schema for Auto-sklearn}\label{algo:1}
\end{algorithm}
%The detector sequentially receives feedback from the real classification which is assumed to be obtained after the prediction is made, when drift is detected, it erases previous evidence. Different adaptation methods were tested and are described below.

\subsubsection{Drift detector}
In a strict LML - AutoML context, where no human intervention is expected, selecting a drift detection algorithm is not trivial, different methods perform differently depending on the type of drift in the data: incremental, abrupt, reoccurring, gradual. Some of these methods also require parameter tuning to achieve better results. For this experiment, we decided to use a particular drift detector with default hyper-parameters to make the problem more manageable and postpone to future work the automatic selection of this type of models.  Specifically, %Among the available options, 
the Fast Hoeffding Drift Detection Method (FHDDM) method was selected because according to experiments in~\citep{tornado} it was found that in many datasets its performance can be compared to the state-of-the-art. 
%\todo[author=HJ,inline]{@Jorge: clarify whether you did these experiments or are you referring to the FHDDM paper}
%Default parameters were considered. % for such method, see The considered parameters being n = 100 and $\delta$ = 0.000001. Where n represents the size of a window that slides on the classification results and $\delta$ a parameter for the test used in the method to detect deviation. Specifically, $\delta$ is the desired probability of error allowed between the empirical mean and the true mean of a set of random variables n, according to Hoeffding's inequality this difference is at least $\epsilon_H$ where $\epsilon_H = \sqrt[]{\frac{1}{2n}ln\frac{2}{\delta}}$. 
%\todo[author=HJ,inline]{@Jorge: please add a few words on what n and $\delta$ are.}
The implementation of FHDDM in  Tornado~\citep{tornado} was integrated with AutoSKlearn.
%The detection method has an important impact in the performance of the proposed method so it is essential to select the an adequate one, this work is limited to experimenting with FHDDM but automatically choosing the best detector given the dataset would be ideal.

\subsubsection{Model adaptation methods}
Three approaches were implemented for adapting the model generated by Auto-Sklearn, a global replacement strategy where the initial model found by Auto-Sklearn is retrained with new examples and two variants of model management strategies.

\noindent \textbf{Model replacement.} A traditional concept drift adaptation where the model is globally replaced with a new one. Taking advantage of the Auto-Sklearn improvements, the model is not trained from scratch. Having the meta-features learned with the meta-learning process from Auto-Sklearn, these are used for the creation of  the new model, which is a new ensemble with the best models found by the SMAC process, the learning for the experimentation was performed storing all data received instead of only the current batch. 

\noindent \textbf{Model management.} Two variants of this formulation were considered. The first strategy assumes the data is generated by a mixture distribution (that formed by previous batches and the current one) and the ensemble weights from the initially learned ensemble are \textit{updated} with new data, either: (1) using the latest batch or (2) using all the stored data up to the current batch, we call these variants WU-latest and WU-all, respectively. Creators of AutoSKlearn~\citep{autosklearn}  found that  uniformly weighted ensembles did not work well. Instead, a greedy \textit{ensemble selection} algorithm was used to build the ensemble. Such same algorithm is used in this work to update the weights of each model under the model management variants. With the new data every model performance is computed again and the ensemble is rebuild with new information. 
%\todo[author=HJ,inline]{@Jorge: It is not clear to me whether the following refers to the three heuristics, or to the two model management variants, please clarify}

In a second model management variant called \emph{Add New}, new models are learned with all the stored data, meta-learning process is also skipped in this case, and using auto-sklearn ensemble construction methods these new models are integrated in the initial ensemble. However, the implementation of this last strategy did not work as expected since the original design of auto-sklearn does not consider adding new models to the ensemble so a new model selection algorithm for the ensemble construction ought to be implemented.

\section{Experiments and results}
\label{sec:experiments}
%This section presents experimental results obtained with the methods described in Section~\ref{sec:automlll}. 
\subsection{Data}
For the empirical evaluation we used benchmark data form both concept drift and the AutoML fields, where the AutoML data sets considered are known to incorporate temporal dependencies across instances. The considered data sets are described in Table~\ref{tab:datasets}.
\begin{table}[htbp]\small{
\floatconts
  {tab:datasets}%
  {\caption{Datasets considered for experimentation.}}%
  {\begin{tabular}{llll}\hline 
  %\bfseries Dataset & \bfseries Result\\  
  \multicolumn{4}{c}{\textbf{Concept drift datasets}}\\
  \bfseries Dataset & \bfseries  instances	& \bfseries attributes	& \bfseries Reference\\\hline
Chess&	503&	8&	\citep{vzliobaite2010change}\\
Poker&	100,000&	10&	\citep{olorunnimbe2015intelligent}\\
Electricity&	45,312&	8& \citep{baena2006early}\\
Stagger&	70,000&	3&	\citep{gama}\\\hline
\multicolumn{4}{c}{\textbf{AutoML2 challenge data sets}}\\
\bfseries Dataset & \bfseries  instances	& \bfseries attributes	& \bfseries Reference\\\hline
PM&	49,964&	89&	\citep{guyon2018}\\
RH&	60,042&	76&	\citep{guyon2018}\\
RI&	57,306&	113&	\citep{guyon2018}\\
RL&	56,209&	22&	\citep{guyon2018}\\
RM&	55,239&	89&	\citep{guyon2018}\\\hline
  \end{tabular}}}
\end{table}
%for both the evaluation of concept drift and the 
%\todo[author=WW,inline]{Feel free to modify this paragraph}
%REC dataset was collected from a content recommender system, sensitive information was removed in this dataset. The target is to predict whether a user will be interested in a content at a time, this is measured by whether the user finished reading/listening/watching the content. The data was collected day by day, with about 30 features including non-sensitive user features and content features. In this dataset, we gathered 21 days of data, there are up to 3 million instances in total.  Here are basic observations of the drifting essential of the dataset: (1) Positive instance ratios differ at most 8\% in these days. (2) There are about 30\% users are new to the recommender system each day. (3) There are about 1\% items are new to the recommender system each day. Moreover, in this dataset, there are some varied length features, e.g. tags of the content, words in the title of the content and so on. Time features are included in this dataset, one example of these features is the publishing time of the content. There are also regenerated unique identifiers for users and items, which have a large number of values following a power law. This dataset is a typical example of many real-world applications, e.g. recommender system, on-line advertising, fraud detection, customer relationship management and so on.

Concept drift data sets have been widely used for the evaluation of drift detection techniques. %In addition, the methods were tested in concept drift benchmark datasets, 3 real-world and 1 synthetic dataset were used. These are briefly described below.
\emph{Chess} comprises 2 years of chess games collected from \url{chess.com}, %the data 
%contains game records from two years, 
the task is to predict whether a player will win or lose a match given players and game setup features, the player increases its abilities through time and faces more skilled players, this is where the concept drift is expected. In the \emph{Poker} dataset  
%100.000 instances of poker dataset were also used, 
each instance is an example of a hand having five playing cards drawn from a standard deck. Each card is described by two attributes (suit and rank) for a total of ten  attributes. The class predicts the poker hand, 10 classes are considered. Since the poker hands were generated in order (varying suits and ranks in order), the concept drifts multiple times.  %; the data set is extremely imbalanced, the distribution is dominated by two classes, and increasing the data sample increases the chances of %two out of the 10 classes comprise more than 90\% of the samples, hence as more data is available, samples for new classes may appear. %. , with few instances a classifier cannot infer poker rules so the drift happens as more examples arrive. 
%this dataset has been extensively used to evaluate concept drift.  
The \emph{electricity} market dataset was collected from the Australian New South Wales Electricity Market. % The prices in this market are affected by demand and supply, 
Each instance represents the market state in a period of 30 minutes. The task in this dataset is to predict if the market prices are going up or down for the next period of time. Finally, a synthetic Stagger dataset was generated to test the methods, three drift points are artificially induced were the concept changes abruptly. According to previous work,  we used accuracy as evaluation measure for these datasets. 

AutoML data sets comprise undisclosed data from click user data, these datasets were used in the AutoML2 challenge~\citep{guyon2018}. Although the data has been keep confidential, it is known that there is a temporal dependency across instances. We expect the proposed methods to capture these dependencies and improve the performance of a straightforward AutoML technique. We know \emph{RH} and \emph{RI} were collected from applications in startup phase, and behavior changed dramatically during that time. \emph{RL} was collected from content recommendation applications, the items were user generated contents, there were a certain proportion of new contents generated each day, the user preferences of the items were changed along time. \emph{PM} and \emph{RM} were collected from a mature recommender system in different time periods,  the items were cars, the candidate set of cars was relatively stable along time, the user preferences were also relatively stable compared to other datasets. For these datasets we used the normalized area under the ROC curve as evaluation measure. 

\subsection{Results}
For each of the considered datasets the following setup was considered. Training partitions were used to obtain an AutoSKlearn model. For concept drift datasets, the data was divided into batches of equal size, and the first batch was used as training set. For the AutoML datasets the predefined training partitions were used for generating the initial model.  After that, the test set was processed in batches. The actual model was used to predict labels for the  batch $t$ and performance was estimated, after that, test labels for this batch were made available to the method. Then we applied the drift detector. If drift was detected, the methods described in Section~\ref{sec:automlll} were applied to adapt the current model, which was then used for batch $t+1$. In experiments, we compare the performance of the initial AutoSKlearn model (\emph{Base}) with the replacement, WU-all, WU-batch and the Add New methods. 

Table~\ref{tab:benchmark} reports the average (across batches) performance for the concept drift datasets.  Regarding the real datasets, in all but one configuration (\emph{Poker} dataset with WU-latest method) the drift aware AutoML solutions outperformed the base model. As it can be expected, replacing the initial model with a new model obtained by applying AutoSKlearn with all available data obtained the best performance. Model replacement, however, is the most computationally expensive solution. A competitive alternative is the WU-all method, in which ensemble weights are updated by using all of the available data. In our experiments, WU-all method was in the worse case about 5 times faster than the replacement technique. The speedup was more noticeable in large datasets. 

Interestingly, most drift aware methods failed with the synthetic data set (\emph{Stagger}), this was due in part to the abrupt drift present in data: synthetic drift is simulated by inverting the concept (negative instances are considered as positive ones). 
%this is due to the abrupt concept drift in the data, before the drift, a specific value for each attribute determines the positive class having the rest of the instances without this values as negatives, after the drift the "rule" changes and the values that determined a positive class now indicate a negative class, a new set of values for each attribute are the new rule for positive instances. 
WU-all and WU-latest models only consider models built before the drift (i.e. with the rule before the drift) and thus fail to classify correctly after  drift detected. The Add new method also tries to incrementally adjust to changes, but AutoSKlearn fails to find an ensemble that classifies correctly both instances before and after the drift. %, since it finds that models only predict at random it keeps predicting with the first model. 
On the other hand, the replacement method builds models from scratch after drift is detected and is able to improve the accuracy of the ensemble.
\begin{table}[htbp]
 % The first argument is the label.
 % The caption goes in the second argument, and the table contents
 % go in the third argument.
\floatconts
  {tab:benchmark}%
  {\caption{Results on benchmark data.}}%
  {\begin{tabular}{lllll|l}
  %\bfseries Dataset & \bfseries Result\\
  \bfseries Method & \bfseries  Electricity	& \bfseries Poker	& \bfseries Chess	& \bfseries Stagger & \bfseries Rank\\\hline
Base& 67.15&	67.97&	38.23&	54.09&4.5\\
Replacement& \textbf{76.44}&	\textbf{90.38}&	\textbf{58.13}&	\textbf{78.81}&1\\
WU-all& 70.23&	76.89&	52.62&	54.09&2.75\\
WU-latest& 67.95&	67.49&	53.24&	54.09&3.5\\
Add new model& 67.47&	74.98&	47.28&	54.14&3.25\\\hline
  \end{tabular}}
\end{table}

Figure~\ref{fig:B} shows the per-batch improvement of AutoML drift aware methods vs. using the initial model. The figure shows a somewhat erratic behavior at times, although, in general, these plots confirm the results presented in Table~\ref{tab:benchmark}. 
It can be seen from these plots that the WU-batch method is very sensitive to the batch used to adjust weights (see rightmost top plot). The Replacement and WU-all strategy consistently outperform the base model in the real data sets, and for the synthetic dataset (rightmost bottom) all methods but replacement perform poorly.  
\begin{figure}[h!tb]
 % Caption and label go in the first argument and the figure contents
 % go in the second argument
\floatconts 
  {fig:B}
  {\caption{Performance of drift aware AutoML variants. From top to bottom and left to right results for: Electricity, Poker, Chess and Stagger are shown, respectively.}}
  {\includegraphics[width=0.4\linewidth]{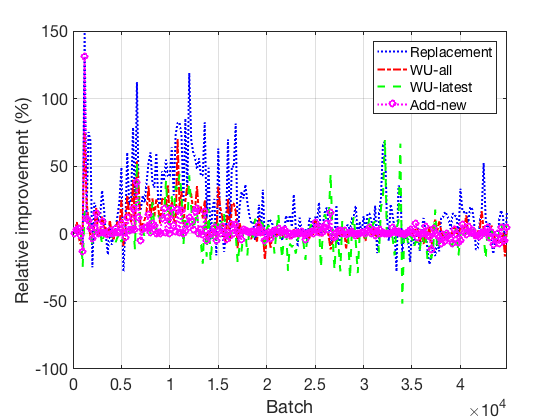}
  \includegraphics[width=0.4\linewidth]{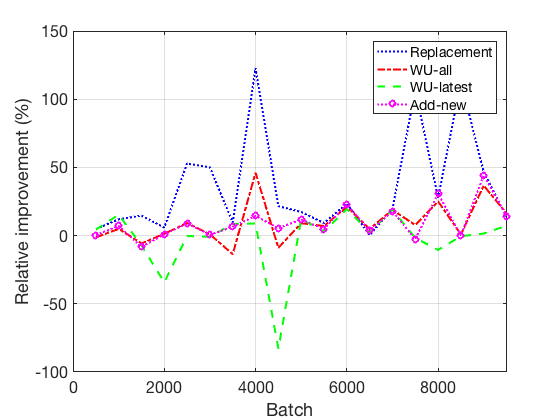}
  \includegraphics[width=0.4\linewidth]{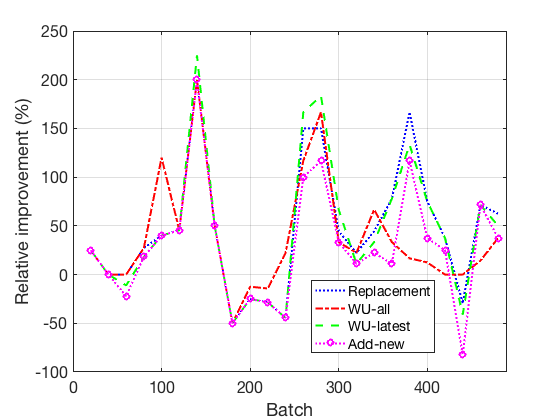}
  \includegraphics[width=0.4\linewidth]{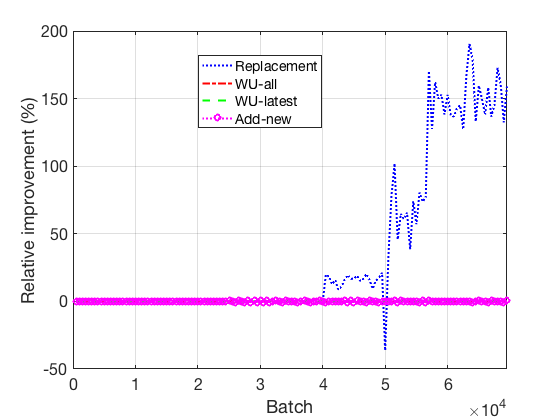}}
\end{figure}

Table~\ref{tab:automl2} shows the results obtained in the AutoML2 challenge data sets, performance is reported in terms of the normalized AUC. For \emph{PM} and \emph{RM} datasets the drift detector did not detect drift in any batch, and therefore the base model was used for making predictions in all the batches. This could be due to the maturity of the application from which \emph{PM} and \emph{RM} were collected from. 
%This could be due to the presence of the \emph{light} 
The drift of concept  evolves slowly and it cannot be captured by the batch size. On the other hand, for datasets collected from startup businesses or UGC recommender systems, due to the changing nature of the applications, for \emph{RH, RI} and \emph{RL}, at least one of the adaptive AutoML methods outperformed considerably the base performance. For \emph{RL}, the Replacement method improved relatively the performance of the base model by 47\%. For \emph{RI} the WU-latest method improved the initial performance by more than 60\%. Whereas for \emph{RH} all of the adaptive AutoML variants outperformed the reference performance; the highest relative improvement being of 73\% obtained by the WU-all strategy. 
\begin{table}[htbp]
 % The first argument is the label.
 % The caption goes in the second argument, and the table contents
 % go in the third argument.
\floatconts
  {tab:automl2}%
  {\caption{Results on data from the AutoML2 challenge.}}%
  {\begin{tabular}{llllll}
  %\bfseries Dataset & \bfseries Result\\
  \bfseries Method & \bfseries  PM$^*$	& \bfseries RH	& \bfseries RI	& \bfseries RL& \bfseries  RM$^*$\\\hline
Base&	0.433&	0.192&	0.299&	0.340&	0.264\\
Replacement&	0.433&	0.197&	0.092&	\textbf{0.478}&	0.264\\
WU-all&	0.433&	\textbf{0.370}&	0.199&	0.212&	0.264\\
WU-latest&	0.433&	0.270&	\textbf{0.450}&	0.405&	0.264\\
Add new&	0.433&	0.298&	0.184&	0.277&	0.264\\\hline
  \end{tabular}}
\end{table}

Although more experiments are needed, the %inconsistency of 
results in AutoML2 data suggest/confirm that each dataset presents a different variant of the  concept drift phenomenon.  \emph{RL} seems to follow the behavior of datasets in Table~\ref{tab:benchmark}. \emph{RH} seems to exhibit a slowly progressing drift, where WU-all takes advantage. While \emph{RI} seems to show a drift aligned with the concept drift detector (hence the WU-latest method is able to find quite useful weights). %Yet, these are preliminary conclusions, more experiments and analyses are required to draw more conclusive findings. 
% \begin{figure}[htbp]
%  % Caption and label go in the first argument and the figure contents
%  % go in the second argument
% \floatconts
%   {fig:A}
%   {\caption{Drift direct comparison}}
%   {\includegraphics[width=0.49\linewidth]{images/FigAEL}
%   \includegraphics[width=0.49\linewidth]{images/FigAPo}
%   \includegraphics[width=0.49\linewidth]{images/FigACs}
%   \includegraphics[width=0.49\linewidth]{images/FigASt}}
% \end{figure}
%\textcolor{red}{Wei Wei, Yuan }
%\textcolor{red}{All, lead by Jorge, Eduardo, Hugo}
\section{Conclusions}
\label{sec:conlcusions}
%\textcolor{red}{All, lead by Jorge, Eduardo, Hugo}
AutoML is an increasingly trending topic because of its relevance given the amount of data being generated nowadays. Although very effective AutoML methods have been proposed so far, to the best of our knowledge all of them focus on the assumption of static (iid) data. However, on-line data is constantly changing and therefore, static solutions may not be useful for evolving data streams. This paper described preliminary experimentation on the evaluation of the drift adaptive capabilities of a state of the art AutoML solution. To the best of our knowledge this is the first work dealing with AutoML in the presence of drift. Experimental results in benchmark data confirm the usefulness of the proposed mechanisms for dealing with drift. Although our results are far from being conclusive, they bring some light into the performance of AutoML for evolving data streams. 
\bibliography{jmlr-sample}

\end{document}